\documentclass{article}
\usepackage{spconf,amsmath,amssymb,graphicx}
\usepackage{booktabs,multirow,cite,hyperref,xcolor,tikz}
\usetikzlibrary{arrows.meta,positioning,fit,calc}
\hypersetup{hidelinks}

\definecolor{signalblue}{RGB}{57,106,177}
\definecolor{signalorange}{RGB}{218,124,48}
\definecolor{signalgreen}{RGB}{62,150,81}
\definecolor{lightgray}{RGB}{242,244,247}

\newcommand{\GeoTokenAcc}{$.143\pm.003$}
\newcommand{\GeoDecodedDist}{$.393\pm.007$}
\newcommand{\BaseTokenAcc}{$.133\pm.004$}
\newcommand{\BaseDecodedDist}{$.606\pm.006$}

\title{GEORVQ: DECODER-AWARE GEOMETRY FOR RESIDUAL-TOKEN PREDICTION IN PHYSIOLOGICAL SIGNALS}
\name{Bo Cui \qquad Yaowen Zhang }
\address{Biomedical Signals and Systems, University of Twente, Enschede, The Netherlands \\
\texttt{ m.r.cui@utwente.nl, y.zhang-12@utwente.nl}}

\begin{document}
\ninept
\maketitle

\begin{abstract}
Residual vector quantization (RVQ) turns physiological waveforms into compact token sequences, but conventional masked modeling treats every incorrect token as equally costly. We propose GeoRVQ, a coarse-to-fine masked token model whose objective reflects the local response of a frozen waveform decoder. Decoder-induced costs define geometry-aware soft targets and expected distortion, while quantizer-causal prediction follows residual dependencies from coarse to fine levels. In a descriptive aggregate over MIMIC-IV Waveform, VitalDB, and CODE-15\%, GeoRVQ increases exact token accuracy from $.133\pm.004$ to $.143\pm.003$, reduces decoded distance from $.606\pm.006$ to $.393\pm.007$, and increases R-peak F1 from $.784\pm.004$ to $.837\pm.008$ under matched model and training conditions. Across 45 held-out code substitutions, decoder-induced cost has a Spearman correlation of $.85$ with realized decoded cost, compared with $.54$ for Euclidean codeword distance. These results indicate that decoder-aware objectives can improve waveform and event preservation without requiring a large increase in exact token accuracy.
\end{abstract}


\section{Introduction}
\label{sec:intro}
Neural codecs convert dense waveforms into compact sequences of discrete tokens \cite{oord2017vqvae,defossez2022encodec,kumar2024dac}. For ECG and PPG, this representation shortens long recordings while preserving a decoder that maps the tokens back to waveforms \cite{avramidis2025biocodec}. Our previous work introduced a hierarchical RVQ tokenizer for compact cross-modal ECG--PPG translation and cross-frequency physiological signal synthesis \cite{cui2026clmt}. The resulting tokens can then be processed by masked models to recover missing signal segments or modalities \cite{mizrahi2023fourm}.

However, standard masked-token prediction overlooks two properties of residual vector quantization (RVQ)~\cite{defossez2022high}.
First, token errors do not have equal effects after decoding.
Replacing a token with a similar codeword may barely change the waveform, whereas another substitution may shift or remove an R peak.
Cross-entropy treats these two errors equally because it considers only whether the predicted token ID is correct~\cite{mizrahi20234m, avramidis2025neural}.
Second, RVQ levels have an inherent coarse-to-fine structure~\cite{kumar2023high}.
Early levels capture the main waveform trajectory, while deeper levels refine the remaining residual.
Predicting every level in parallel may therefore estimate fine details before the coarse signal has been resolved~\cite{copet2024simple}.

We propose GeoRVQ to account for both properties. A frozen pretrained codec provides the target tokens and measures the waveform change caused by each codeword substitution. GeoRVQ uses these decoder-induced costs to construct soft targets and penalize predictions according to their decoded effects. Its quantizer-causal heads first predict the coarse RVQ level and then condition deeper-level predictions on the resolved coarse tokens. Temporal context remains bidirectional throughout this process.

We test whether decoder-aware training improves decoded waveform and event preservation when exact token accuracy changes only slightly. Comparisons include one-hot cross-entropy, label smoothing, Euclidean codeword geometry, and decoder-aware objectives under the same codec, backbone, and RVQ depth. We separately evaluate the decoder-induced geometry and the contribution of coarse-to-fine prediction. Adaptive rate routing is excluded to keep the evaluation focused on these two components.

\section{GeoRVQ}
\label{sec:method}

\subsection{RVQ token prediction}

\begin{figure*}[t]
  \centering
  \includegraphics[width=0.98\textwidth]{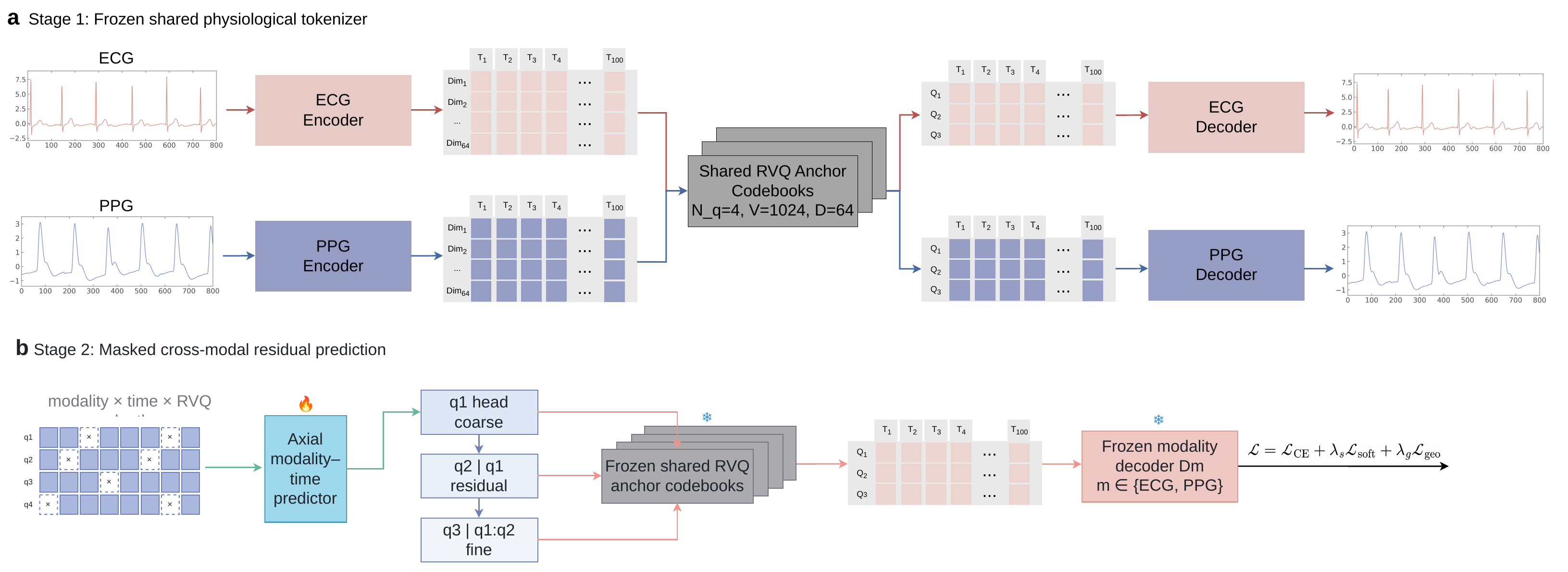}
  \caption{\textbf{GeoRVQ architecture.} A frozen shared physiological codec converts ECG and PPG into hierarchical RVQ tokens. The codec follows the hierarchical RVQ design introduced in CLMT \cite{cui2026clmt}, while the present work focuses on decoder-aware token prediction rather than codec design. The trainable axial predictor processes a masked modality--time--depth tensor. Quantizer-causal heads resolve the coarse code $q_1$, then condition deeper levels on previously resolved codes. Predicted codes are mapped through the frozen shared codebooks and modality decoder. Decoder-induced geometry is precomputed from training contexts and enters only through the soft-target and expected-cost terms. The tensor drawing is schematic; the reported fixed-depth experiments evaluate $q_1$, $q_{1:2}$, and $q_{1:3}$.}
  \label{fig:architecture}
\end{figure*}

Figure~\ref{fig:architecture} separates the frozen physiological codec from the trainable masked predictor. Let a frozen encoder convert a waveform $x^m$ from modality $m\in\{e,p\}$ into latent frames $z_{1:T}^m$. A shared $L$-level RVQ represents each frame by indices $k_{t,1:L}^m$, with reconstruction
\begin{equation}
 \hat z_t^m=\sum_{\ell=1}^{L}c_{\ell,k_{t,\ell}^m}, \qquad
 \hat x^m=D_m(\hat z^m),
\end{equation}
where $c_{\ell,j}$ is shared codeword $j$ at level $\ell$ and $D_m$ is the frozen modality decoder. A mask specifies the observed modality--time--level entries and the targets. An axial backbone first exchanges temporal and cross-modal context. Quantizer-causal heads factorize
\begin{equation}
p(k_{t,1:L}\mid\mathcal{O})=
p(k_{t,1}\mid\mathcal{O})\prod_{\ell=2}^{L}
p(k_{t,\ell}\mid k_{t,<\ell},\mathcal{O}),
\end{equation}
without imposing an autoregressive order across time. During training, scheduled sampling exposes fine-level heads to predicted rather than exclusively teacher-forced coarse tokens.

\subsection{Decoder-induced code geometry}

Euclidean codeword distance does not account for decoder anisotropy or residual interactions. For a codeword pair $(i,j)$ at level $\ell$, we estimate a context-averaged decoded cost
\begin{equation}
 d_{\ell}(i,j)=\mathbb{E}_{u\sim\mathcal{C}_{\ell,i}}
 \left[\rho\!\left(D_m(u\oplus c_{\ell,i}),D_m(u\oplus c_{\ell,j})\right)\right],
\end{equation}
where $u$ contains the remaining latent context and $\oplus$ replaces one residual code. The cost $\rho=w_xE_x+w_{\nabla}E_{\nabla}+w_eE_e$ combines normalized waveform, derivative, and event-neighborhood errors, with fixed non-negative weights that sum to one. Contexts $\mathcal{C}_{\ell,i}$ are sampled only from the training split. We compute the full cost row for the expected-cost term and retain the support $\mathcal{S}_{\ell,i}$ containing the source code and its $K$ nearest neighbors for the soft target:
\begin{equation}
 y_{\ell,i}(j)=
 \begin{cases}
 \displaystyle\frac{\exp[-d_{\ell}(i,j)/\tau]}
 {\sum_{r\in\mathcal{S}_{\ell,i}}\exp[-d_{\ell}(i,r)/\tau]},
 & j\in\mathcal{S}_{\ell,i},\\[4pt]
 0,&\text{otherwise.}
 \end{cases}
\end{equation}
The geometry term is the expected decoded cost under predicted probabilities $p_{\ell}$:
\begin{equation}
 \mathcal{L}_{\mathrm{geo}}=\sum_{t,\ell}\sum_j
 p_{t,\ell}(j)d_{\ell}(k_{t,\ell},j).
\end{equation}
The full objective combines hard-label cross-entropy, soft-target cross-entropy, and expected cost:
\begin{equation}
 \mathcal{L}=\mathcal{L}_{\mathrm{CE}}+\lambda_s\mathcal{L}_{\mathrm{soft}}
 +\lambda_g\mathcal{L}_{\mathrm{geo}}.
\end{equation}
The hard target anchors code identity; the two geometry terms are ablated separately to distinguish useful structure from generic smoothing.

\section{Experiments}
\label{sec:experiments}

\subsection{Datasets, splits, and preprocessing}

Pretraining uses MIMIC-IV Waveform v0.1.0, VitalDB, and CODE-15\% \cite{moody2022mimicwaveform,lee2022vitaldb,ribeiro2021code15}. MIMIC-IV and VitalDB contribute paired ECG--PPG windows after channel-level quality control; CODE-15\% contributes ECG only and is excluded from paired translation. Dataset-balanced sampling prevents the larger ECG-only source from dominating training.

For MIMIC-IV, we use its 200 records from 198 subjects and fix a subject-disjoint 138/20/40 train/validation/test split. Codec and masked-model stages use the same split; the 40 test subjects are excluded from all training, validation, graph construction, hyperparameter tuning, and checkpoint selection. Each observation is an 8-s paired ECG--PPG window resampled to 100 Hz under fixed preprocessing. A 25 Hz ECG input is generated by predefined low-pass anti-alias filtering followed by downsampling, not sample slicing. Adjacent or overlapping windows never cross splits, and no evaluation uses post-hoc temporal alignment.

VitalDB uses a case-disjoint 70/10/20\% train/validation/test partition after the same frozen paired-signal quality-control rules. CODE-15\% uses a patient-disjoint 70/10/20\% partition, stratified by diagnostic labels where available; lead I is used for the single-lead experiments. The split is fixed before window generation and shared by the codec, masked predictor, baselines, and ablations. All normalization statistics and decoder-geometry contexts are fitted on the training partition of each dataset and then frozen.

\subsection{Comparisons and tasks}

The primary loss comparison holds the codec, mask schedule, backbone, parameter count, update budget, and active RVQ depth fixed. We compare one-hot cross-entropy, label smoothing, Euclidean codeword soft targets, decoder-aware soft targets, expected decoder cost alone, and the full objective. Parallel and quantizer-causal heads are compared under the same backbone. Fixed $q_1$, $q_{1:2}$, and $q_{1:3}$ operating points define the depth axis; no learned rate router is used.

Tasks comprise random and contiguous-block inpainting, paired ECG--PPG translation on MIMIC-IV and VitalDB, and 25-to-100/250/500 Hz ECG synthesis where supported. CODE-15\% is used only for ECG-compatible reconstruction, cross-rate synthesis, and event preservation. The primary analysis is masked token prediction followed by decoding; downstream diagnosis is secondary and cannot substitute for the geometry test.

\subsection{Metrics and statistics}

We report exact token accuracy and negative log-likelihood over masked targets, decoded waveform metrics, power spectral density (PSD) error, and event-level measures where eligible. The same detector and matching rule are applied to every method. Metrics are first averaged within each subject or case, then within each eligible dataset and task. Dataset-level summaries receive equal weight in the aggregate comparison, which prevents larger cohorts from dominating the reported value. Main models and decision-critical ablations use three independent training runs. Displayed uncertainty is the across-run sample standard deviation and is interpreted descriptively; it does not represent population-level subject uncertainty. No window is treated as an independent replicate. Geometry validity is reported as Spearman correlation over 45 substitutions, with 15 substitutions from each evaluated RVQ level.

\section{Results}
\label{sec:results}

\subsection{Decoded fidelity improves more than token accuracy}

\begin{table*}[t]
\centering
\caption{Matched comparison of token objectives and RVQ factorization. All methods use the same frozen codec, mask schedule, backbone, training budget, and active RVQ depth. Metrics are macro-averaged across eligible dataset summaries, with equal dataset weight. Values are descriptive mean$\pm$SD over three independent runs.}
\label{tab:geo-main}
\setlength{\tabcolsep}{4.5pt}
\resizebox{\textwidth}{!}{%
\begin{tabular}{lcccccc}
\toprule
Method & Token acc. $\uparrow$ & NLL $\downarrow$ & Decoded dist. $\downarrow$ & PCC $\uparrow$ & R-F1 $\uparrow$ & PSD error $\downarrow$ \\
\midrule
One-hot cross-entropy & $.133\pm.004$ & $4.631\pm.036$ & $.606\pm.006$ & $.969\pm.003$ & $.784\pm.004$ & $.083\pm.004$ \\
Label smoothing & $.129\pm.002$ & $4.645\pm.062$ & $.558\pm.006$ & $.971\pm.001$ & $.792\pm.010$ & $.078\pm.003$ \\
Euclidean codeword geometry & $.133\pm.003$ & $4.572\pm.011$ & $.516\pm.015$ & $.974\pm.001$ & $.808\pm.011$ & $.068\pm.004$ \\
Decoder geometry, parallel heads & $.136\pm.001$ & $4.555\pm.018$ & $.441\pm.015$ & $.984\pm.001$ & $.817\pm.003$ & $.053\pm.003$ \\
Decoder geometry, no soft targets & $.138\pm.002$ & $4.523\pm.021$ & $.425\pm.010$ & $.986\pm.001$ & $.828\pm.004$ & $.048\pm.003$ \\
\textbf{GeoRVQ} & $\mathbf{.143\pm.003}$ & $\mathbf{4.456\pm.048}$ & $\mathbf{.393\pm.007}$ & $\mathbf{.989\pm.001}$ & $\mathbf{.837\pm.008}$ & $\mathbf{.041\pm.002}$ \\
\bottomrule
\end{tabular}}
\end{table*}

Under matched model, training, and active-depth conditions, GeoRVQ achieved exact token accuracy \GeoTokenAcc\ and decoded distance \GeoDecodedDist, compared with \BaseTokenAcc\ and \BaseDecodedDist\ for one-hot cross-entropy (Table~\ref{tab:geo-main}). The $.010$ absolute accuracy change accompanied a $35.2\%$ reduction in decoded distance. PCC increased from $.969\pm.003$ to $.989\pm.001$, R-peak F1 increased from $.784\pm.004$ to $.837\pm.008$, and PSD error decreased from $.083\pm.004$ to $.041\pm.002$. Label smoothing changed decoded metrics without increasing exact accuracy, whereas both decoder-aware objectives produced larger descriptive gains.

\begin{figure*}[!t]
  \centering
  \IfFileExists{georvq_results.pdf}{%
    \includegraphics[width=0.98\textwidth]{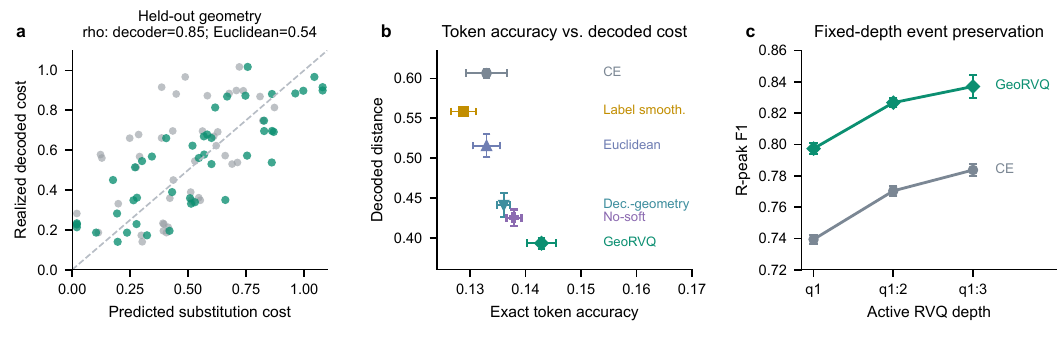}%
  }{%
    \fbox{\parbox[c][50mm][c]{0.95\textwidth}{\centering
    \textbf{RESULT FIGURE PLACEHOLDER}\\[2pt]
    (a) graph distance vs. held-out distortion;\\
    (b) token accuracy vs. decoded error;\\
    (c) RVQ depth versus event preservation.}}
  }
  \caption{\textbf{Decoder geometry and decoded signal quality.} (a) Predicted substitution cost versus realized decoded cost in contexts excluded from graph construction. Each point is one of 45 substitutions averaged across held-out contexts. Spearman $\rho$ is $.85$ for decoder-induced cost and $.54$ for Euclidean codeword distance. (b) Exact token accuracy versus decoded distance for matched objectives. (c) R-peak F1 at fixed $q_1$, $q_{1:2}$, and $q_{1:3}$ depths. Error bars in (b,c) are sample SD over three independent runs.}
  \label{fig:results}
\end{figure*}

\subsection{The decoder graph predicts held-out consequences}

Figure~\ref{fig:results}(a) evaluates the geometry independently of the masked predictor on the sampled substitutions. Decoder-induced cost reached a Spearman correlation of $.85$ with realized waveform cost, whereas Euclidean codeword distance reached $.54$. This descriptive comparison is consistent with transfer of the context-averaged geometry to unseen latent neighborhoods. Figure~\ref{fig:results}(b) provides the complementary model-level result: methods with similar token accuracy occupy different decoded-distance regimes.

\subsection{Geometry and quantizer causality are complementary}

\begin{table}[t]
\centering
\caption{Mechanism ablation at the $q_{1:3}$ operating point. Values are descriptive mean$\pm$SD over three independent runs.}
\label{tab:geo-ablation}
\setlength{\tabcolsep}{3.0pt}
\resizebox{\columnwidth}{!}{%
\begin{tabular}{lcccc}
\toprule
Variant & Token acc. $\uparrow$ & Dec. dist. $\downarrow$ & R-F1 $\uparrow$ & R-time MAE (ms) $\downarrow$ \\
\midrule
No decoder geometry & $.137\pm.002$ & $.502\pm.012$ & $.807\pm.004$ & $10.4\pm0.4$ \\
No quantizer causality & $.136\pm.001$ & $.441\pm.015$ & $.817\pm.003$ & $9.7\pm0.1$ \\
No soft targets & $.138\pm.002$ & $.425\pm.010$ & $.828\pm.004$ & $9.4\pm0.2$ \\
\textbf{Full GeoRVQ} & $\mathbf{.143\pm.003}$ & $\mathbf{.393\pm.007}$ & $\mathbf{.837\pm.008}$ & $\mathbf{9.2\pm0.2}$ \\
\bottomrule
\end{tabular}}
\end{table}

Removing decoder geometry increased decoded distance from $.393\pm.007$ to $.502\pm.012$ and reduced R-peak F1 from $.837\pm.008$ to $.807\pm.004$ (Table~\ref{tab:geo-ablation}). Removing quantizer-causal conditioning increased decoded distance to $.441\pm.015$, reduced R-peak F1 by $.020$, and increased R-peak timing MAE from $9.2\pm0.2$ to $9.7\pm0.1$ ms. Expected decoder cost without soft targets retained part of the improvement. These descriptive ablations associate the largest distance change with decoder geometry and a smaller event-timing change with coarse-to-fine conditioning.

\subsection{Fixed-depth and cross-dataset consistency}

The depth--event comparison in Fig.~\ref{fig:results}(c) favors GeoRVQ at each evaluated RVQ depth. At $q_1$, $q_{1:2}$, and $q_{1:3}$, GeoRVQ reached R-peak F1 values of $.797\pm.004$, $.827\pm.003$, and $.837\pm.007$, compared with $.739\pm.003$, $.770\pm.003$, and $.784\pm.004$ for one-hot cross-entropy. Dataset-specific decoded distance was $.392$ versus $.614$ on MIMIC-IV, $.407$ versus $.628$ on VitalDB, and $.381$ versus $.576$ on CODE-15\%. The corresponding R-peak F1 values were $.839$ versus $.788$, $.832$ versus $.779$, and $.838$ versus $.784$. These dataset-level summaries show the same descriptive direction, but no population-level inference is claimed from the three-run SD.

\section{Conclusion}

GeoRVQ aligns masked RVQ-token learning with the decoded consequences of token substitutions and follows the coarse-to-fine dependency of residual codes. Across the evaluated cohorts, its descriptive improvements in decoded distance and R-peak preservation were larger than its change in exact token accuracy. Held-out substitution analysis was consistent with decoder-induced cost providing a more informative local geometry than Euclidean codeword distance on the sampled pairs. The context-averaged graph remains an approximation to a nonlinear decoder and may miss rare state-dependent substitutions.

\bibliographystyle{IEEEbib}
\bibliography{references}

\begin{thebibliography}{10}

\bibitem{oord2017vqvae}
Aaron van~den Oord, Oriol Vinyals, and Koray Kavukcuoglu,
\newblock ``Neural discrete representation learning,''
\newblock in {\em Proceedings of the 31st International Conference on Neural Information Processing Systems}, Red Hook, NY, USA, 2017, NIPS'17, p. 6309–6318, Curran Associates Inc.

\bibitem{defossez2022encodec}
Alexandre D{\'e}fossez, Jade Copet, Gabriel Synnaeve, and Yossi Adi,
\newblock ``High fidelity neural audio compression,''
\newblock {\em Transactions on Machine Learning Research}, 2023.

\bibitem{kumar2024dac}
Rithesh Kumar, Prem Seetharaman, Alejandro Luebs, Ishaan Kumar, and Kundan Kumar,
\newblock ``High-fidelity audio compression with improved {RVQGAN},''
\newblock in {\em Advances in Neural Information Processing Systems}, 2023.

\bibitem{avramidis2025biocodec}
Kleanthis Avramidis, Tiantian Feng, Woojae Jeong, Jihwan Lee, Wenhui Cui, Richard~M. Leahy, and Shrikanth Narayanan,
\newblock ``Neural codecs as biosignal tokenizers,''
\newblock {\em arXiv preprint arXiv:2510.09095}, 2025.

\bibitem{cui2026clmt}
Bo~Cui, Xiaowen Song, Yaowen Zhang, Shunzhe Zhang, B.~J.~F. van Beijnum, Monique Tabak, and Ying Wang,
\newblock ``Compact latent manifold translation: A parameter-efficient foundation model for cross-modal and cross-frequency physiological signal synthesis,''
\newblock {\em arXiv preprint arXiv:2605.13248}, 2026.

\bibitem{mizrahi2023fourm}
David Mizrahi, Roman Bachmann, O{\u{g}}uzhan~Fatih Kar, Teresa Yeo, Mingfei Gao, Afshin Dehghan, and Amir Zamir,
\newblock ``{4M}: Massively multimodal masked modeling,''
\newblock in {\em Advances in Neural Information Processing Systems}, 2023.

\bibitem{defossez2022high}
Alexandre D{\'e}fossez, Jade Copet, Gabriel Synnaeve, and Yossi Adi,
\newblock ``High fidelity neural audio compression,''
\newblock {\em Transactions on Machine Learning Research}, 2023,
\newblock Featured Certification, Reproducibility Certification.

\bibitem{mizrahi20234m}
David Mizrahi, Roman Bachmann, O{\u{g}}uzhan~Fatih Kar, Teresa Yeo, Mingfei Gao, Afshin Dehghan, and Amir Zamir,
\newblock ``4m: Massively multimodal masked modeling,''
\newblock in {\em Advances in Neural Information Processing Systems (NeurIPS)}, 2023.

\bibitem{avramidis2025neural}
Kleanthis Avramidis, Tiantian Feng, Woojae Jeong, Jihwan Lee, Wenhui Cui, Richard~M. Leahy, and Shrikanth Narayanan,
\newblock ``Neural codecs as biosignal tokenizers,''
\newblock {\em arXiv preprint arXiv:2510.09095}, 2025.

\bibitem{kumar2023high}
Rithesh Kumar, Prem Seetharaman, Alejandro Luebs, Ishaan Kumar, and Kundan Kumar,
\newblock ``High-fidelity audio compression with improved rvqgan,''
\newblock in {\em Advances in Neural Information Processing Systems (NeurIPS)}, 2023.

\bibitem{copet2024simple}
Jade Copet, Felix Kreuk, Itai Gat, Tal Remez, David Kant, Gabriel Synnaeve, Yossi Adi, and Alexandre D{\'e}fossez,
\newblock ``Simple and controllable music generation,''
\newblock {\em Advances in Neural Information Processing Systems (NeurIPS)}, 2023.

\bibitem{moody2022mimicwaveform}
Benjamin Moody, Sicheng Hao, Brian Gow, Tom Pollard, Alistair Johnson, and Roger Mark,
\newblock ``{MIMIC-IV Waveform Database},'' PhysioNet, version 0.1.0, 2022.

\bibitem{lee2022vitaldb}
Hyeon~Cheol Lee, Yoon~Ji Park, Seong~Bae Yoon, Seok~Min Yang, Dongnyeok Park, and Chul-Woo Jung,
\newblock ``{VitalDB}, a high-fidelity multi-parameter vital signs database in surgical patients,''
\newblock {\em Scientific Data}, vol. 9, pp. 279, 2022.

\bibitem{ribeiro2021code15}
Ant{\^o}nio~H. Ribeiro, Gabriela M.~M. Paix{\~a}o, Emilly~M. Lima, Manoel Horta~Ribeiro, Marcelo~M. Pinto~Filho, Paulo~R. Gomes, Derick~M. Oliveira, Wagner Meira~Jr., Thomas~B. Sch{\"o}n, and Antonio Luiz~P. Ribeiro,
\newblock ``{CODE-15\%}: A large scale annotated dataset of 12-lead {ECG}s,'' Zenodo, 2021.

\end{thebibliography}

\end{document}